\documentclass[pmlr,twocolumn,10pt]{jmlr} % W&CP article

\usepackage{booktabs}
\usepackage{float}
\usepackage{adjustbox} % Add this line
\usepackage{booktabs}
\usepackage{siunitx}

\usepackage[switch]{lineno}

\newcommand{\equal}[1]{{\hypersetup{linkcolor=black}\thanks{#1}}}

\theorembodyfont{\upshape}
\theoremheaderfont{\scshape}
\theorempostheader{:}
\theoremsep{\newline}

\jmlryear{2024}
\jmlrworkshop{Conference on Health, Inference, and Learning (CHIL) 2024, NY} % W&CP title
\title[FlowCyt]{FlowCyt: A Comparative Study of Deep Learning\titlebreak Approaches for Multi-Class Classification in Flow Cytometry Benchmarking}

\author{
  \Name{Lorenzo Bini} \equal{These authors contributed equally.} 
  \Email{lorenzo.bini@unige.ch}\\
  \addr University of Geneva, Switzerland
  \AND
  \Name{Fatemeh Nassajian Mojarrad} \footnotemark[1] 
  \Email{fatemeh.nassajian@unige.ch}\\
  \addr University of Geneva, Switzerland
  \AND
  \Name{Margarita Liarou} \Email{margarita.liarou@unige.ch}\\
  \addr University of Geneva, Switzerland
  \AND
  \Name{Thomas Matthes} \Email{thomas.matthes@hcuge.ch}\\
  \addr Geneva University Hospital, Switzerland
  \AND
  \Name{Stéphane Marchand-Maillet} \Email{stephane.marchand-maillet@unige.ch}\\
  \addr University of Geneva, Switzerland
}
\begin{document}

\maketitle
%\tableofcontents

\begin{abstract}
This paper presents FlowCyt, the first comprehensive benchmark for multi-class single-cell classification in flow cytometry data. The dataset comprises bone marrow samples from 30 patients, with each cell characterized by twelve markers. Ground truth labels identify five hematological cell types: T lymphocytes, B lymphocytes, Monocytes, Mast cells, and Hematopoietic Stem/Progenitor Cells (HSPCs). Experiments utilize supervised inductive learning and semi-supervised transductive learning on up to 1 million cells per patient. Baseline methods include Gaussian Mixture Models, XGBoost, Random Forests, Deep Neural Networks, and Graph Neural Networks (GNNs). GNNs demonstrate superior performance by exploiting spatial relationships in graph-encoded data. The benchmark allows standardized evaluation of clinically relevant classification tasks, along with exploratory analyses to gain insights into hematological cell phenotypes. This represents the first public flow cytometry benchmark with a richly annotated, heterogeneous dataset. It will empower the development and rigorous assessment of novel methodologies for single-cell analysis.
\end{abstract}

\paragraph*{Data and Code Availability}
This paper uses the flow cytometry dataset provided by our University Hospital, which is available in our GitHub benchmark's repository along with our code\footnote{\href{https://viper.unige.ch/flowcyt}{FlowCyt Webpage}}. 

\paragraph*{Institutional Review Board (IRB)}
This study was reviewed and approved by the Institutional Review Board of our University Hospital. 

\section{Introduction}
\label{sec:intro}
In the constantly changing field of analyzing hematologic cell populations, multi-class classification remains a challenging task, requiring innovative diagnostic strategies capable of understanding the complexities of diverse cellular manifestations. Among various diagnostic tools, flow cytometry has become a key element in clinical practices, providing quick insights into cell populations and enabling the identification and characterization of abnormal cell types. However, the diagnostic potential of flow cytometry is counterbalanced by the formidable challenges posed by the complexity and heterogeneity of hematologic populations, requiring a shift in analytical methodologies. 

We endeavor to offer the first public benchmark for flow cytometry data, providing automated solutions for hematological relevant tasks, such as multi-class single-cell classification. It is noteworthy that this is the first available benchmark for flow cytometry data, equipped with ground truth, baseline models and advanced deep learning methods. This benchmark aims to provide a standardized platform for evaluating the performance of various algorithms and models, facilitating the development of novel, more efficient, and accurate diagnostic strategies.

\subsection{Flow Cytometry}
Flow cytometry (FC) is a laboratory technique used to detect and measure physical and chemical characteristics of population of cells or particles in a solution. In medicine, particularly in hematology and immunology, flow cytometry is mainly applied to characterize and count types of white blood cells in the evaluation of infectious diseases, autoimmune disorders, immunodeficiencies, or in the diagnosis of blood cancers such as leukemias or lymphomas.

Typically, samples obtained from the blood or bone marrow of patients are thereby labeled with antibodies that recognize specific surface proteins expressed by the cell populations present. The antibodies are coupled to fluorochromes so that when they are passed before a laser beam light is absorbed and then emitted in a band of wavelengths. Tens of thousands of cells can be quickly examined by modern flow cytometers in a matter of seconds and the data gathered subsequently processed by a computer. Highly complex data sets are thereby generated with data from up to several million cells and information about the presence or not of between 15 to 50 different cell proteins.
%Flow cytometry   is widely used in both clinical and basic research to characterize biological samples at single-cell resolution with multiple protein markers. Firstly, flow cytometry experts label the cells with antibodies tagged with fluorescent markers and employ a flow cytometer to identify the fluorescent signals while the cells rapidly flow past lasers. 
Despite originating in the 1960s \citep{Fulwyler1965,Gray1975}, the fundamental structure of flow cytometry has undergone minimal alteration. Nevertheless, ongoing enhancements to both flow cytometers and fluorescent dyes have substantially accelerated cell analysis and expanded the range of detectable protein markers. In the 2000s, cytometry by time of flight (CyTOF), also referred to as mass spectrometry, was invented \citep{Bandura2009,Bendall2011}.
CyTOF can identify isotope peaks without notable spectrum overlap with the use of heavy metal isotope coupled antibodies. This enables the simultaneous profiling of over 50 protein markers. 
In this work, we focus on providing and describing flow cytometry data, along with baseline performance of automated analysis technique.

%There exist advanced technologies like single-cell RNA-sequencing (scRNA-seq) and cellular indexing of transcriptomes and epitopes by sequencing (CEIT-seq). However, even if these technologies offer the advantage of characterizing cells with a larger number of measurements, disadvantages remain, e.g. they are limited by the high cost and the relatively low number of cells that can be processed. In contrast, cytometry experiments are cost-effective and can characterize hundreds of samples, while most scRNA-seq experiments are typically limited to fewer than 10 samples. Additionally, cytometry can profile a large number of cells per sample, enabling the identification of rare cell populations. Hence, modern cytometry remains one of the most important tools in immunology research.
 
\subsection{Manual FC Data Analysis } 
Traditional manual gating employs a series of two-dimensional plots for data visualization and employs hierarchical gates to identify cell populations \citep{Salama2022}. It has the advantage that  it can integrates prior knowledge into cytometry data analysis, encompassing the functionality of protein markers and the developmental correlation among cell populations. However, it remains a challenge to analyse the  cytometry data  due to its large cell numbers, high dimensionality, human bias  and heterogeneity between datasets.

Special software has been developped which allow cytometrists to analyse this multi-dimensional data,  to focus on specific cell populations, and to establish the phenotypic profile of a cell or cell population. This manual analysis is time-consuming, laborious, needs a high expertise, but still remains subjective and user-dependent.  

Methods are therefore urgently needed which allow the analysis of this data in an efficient, rapid and precise way.

\subsection{Related Work}
%There already exist some ways around this problem to automate every step of the cytometry data analysis, e.g. 
There already exist numerous mathematical and statistical tools to analyze cytometric multi-dimensional data sets such as
cell population identification \citep{Hu2018,Naim2014,Finak2009,Aghaeepour2011,VanGassen2015,Dorfman2016},
data visualization \citep{Amir2013,Becht2019,Qiu2011},
sample classification \citep{Bruggner2014,VanGassen2016,Arvaniti2017,Hu2020},
batch normalization \citep{Schuyler2019,Gassen2020}, quality control \citep{Monaco2016} and trajectory inference \citep{dai2021cytotree}.
These tools employ a wide range of computational
methods which range from rule-based algorithms to machine
learning models.

%The paper is organized as follows. In Section  
% Our strategy is to  leverage the power of  graph attention network (GAT) in the domain of flow cytometry single cell classification.

% Graph neural networks (GNNs) has recently attracted some attention, see e.g.

% GNNs offer an effective approach for extracting insights from data organized in graph structures. Consequently, GNNs find straightforward application in domains where  the data can be represented as a set of nodes and the prediction depends on the relationships (edges) between the nodes.

% In the last years, the use of  GNNs has influenced many areas 

% We can easily apply our approach to larger settings to detect cell populations in a highly heterogenous mixture, which is a crucial step in the diagnosis field.

\section{Dataset Description}
\label{sec:dataset-description}
In this section, we present an overview of FlowCyt, which is the first benchmark for single-cell multi-class classification using flow cytometry data. It utilizes a curated dataset comprising samples from 30 different patients.

\subsection{Biological Description}
Tabular data, especially in the field of biological data analysis, has a distinct and complicated structure. Each row in the table represents a single cell, which is characterized by several parameters like size, granularity, and fluorescence intensity. These parameters collectively constitute a high-dimensional feature space that captures the intricate biological variations among the cells. This data's inherent structure often contains underlying relationships and dependencies that are not immediately apparent.

In this research, we begin by collecting raw data from the bone marrow of patients using a cytometer. The cytometer assesses the physical and chemical features of cells in a fluid as it passes through a laser beam. The surface molecules are fluorescently labeled and then activated by the laser to emit light at different wavelengths. Detectors collect the emitted light and transform it into electric signals. The raw data obtained from the cytometer are saved in Flow Cytometry Standard (FCS) files. Each FCS file comprises multidimensional data that corresponds to thousands of cells, with each cell characterized by various parameters such as size, granularity, and fluorescence intensity.

Thirty bone marrow (BM) samples were obtained from patients who underwent a flow cytometric analysis for diagnostic purposes. All samples were processed by the Diagnostics laboratory of the University Hospital and no malignant disease was detected. Samples were stained with the following antibodies: CD14=FITC, CD19=PE, CD13=ECD,  CD13=PerCP5.5, CD34=PC7, CD117=APC, CD7=APC700, CD16=APC750, HLA-DR=PB and CD45=KO. Between $250'000$ and $1'000'000$ cells were acquired from each sample on a 10-color Navios cytometer (Beckman\&Coulter) and analyzed manually with the KALUZA software (B\&C).

After the exclusion of cell debris, dead cells, and doublets, cell counts were established for the following five different cell populations: B lymphocytes (CD19pos), T lymphocytes (CD7pos), monocytes (CD14pos and CD33pos), mast cells (CD117 strongly pos), and HSPCs (hematopoietic stem and progenitor population, CD34pos). Therefore, we characterized in our dataset the following populations:
\begin{enumerate}
    \item T Lymphocytes: These are a specific type of white blood cell that plays a pivotal role in the adaptive immune system, which is the response that involves the activation of immune cells to fight infection. They are responsible for directly killing infected host cells, activating other immune cells, producing cytokines, and regulating the immune response.
    
    \item B Lmphocytes: These cells are significant contributors to the adaptive immune system. They are responsible for producing antibodies against antigens, which are substances that the immune system recognizes as foreign. Each mature B cell is programmed to make one specific antibody. When a B cell encounters its triggering antigen, it gives rise to many large cells known as plasma cells, each of which is essentially an antibody factory.
    
    \item Monocytes: These are a type of white blood cell and a part of the innate immune system. They play a vital role in the body's defense against infections and other foreign invaders. Monocytes circulate in the bloodstream, and when they migrate into tissues, they differentiate into macrophages or dendritic cells, which are capable of engulfing and digesting pathogens and apoptotic cells.
    
    \item Mast cells: These are a type of immune cell that plays a crucial role in the body's response to allergies and certain infections. Mast cells are found in most tissues, but especially in areas close to the external environment, such as the skin and mucous membranes. They contain granules filled with potent chemicals, including histamine, which they release in response to contact with an allergen. This release triggers inflammation, which can lead to allergic reactions.
    
    \item Hematopoietic stem and progenitor cells (HSPC): These are a type of stem cell found in the bone marrow and cord blood. They have the unique ability to give rise to all other types of blood cells, including red blood cells, platelets, and all types of white blood cells. This makes them crucial for maintaining the body's blood supply and immune system.

\end{enumerate}
Each sample in the dataset includes ground truth labels indicating the corresponding hematological cell type. Researchers can use these labels to train and evaluate classification models aimed at discerning different cell types. Moreover, the dataset enables exploratory analyses such as clustering and dimensionality reduction, allowing comprehensive investigations into the heterogeneity and dynamics of hematological cell populations.

\subsection{Technical Description}
Table \ref{tab:1} displays the features that we utilized for all the analysis and tasks. All files have been gated by hematologists according to the list of cell types (labels) provided below.
\begin{table}[hbtp]
    \tiny
    \centering
    \floatconts
    {tab:1}
    {\caption{\footnotesize Flow cytometry data markers.}}
    {\begin{tabular}{ccc}
    \toprule
        Class & Marker  & Description \\
        \midrule
       0& FS INT & Forward Scatter (FSC) - Cell’s size \\
        1&SS INT & Side Scatter (SSC) - Cell’s granularity\\
         2&CD14-FITC & Cluster of Differentiation 14 - Antigen\\
        3&CD19-PE & Cluster of Differentiation 19 - Antigen\\
        4&CD13-ECD &Cluster of Differentiation 13 - Antigen \\
        5&CD33-PC5.5 & Cluster of Differentiation 33 - Antigen\\
        6&CD34-PC7 & Cluster of Differentiation 34 - Antigen\\
        7&CD117-APC &Cluster of Differentiation 117 - Antigen \\
        8&CD7-APC700 &Cluster of Differentiation 7 - Antigen \\
        9&CD16-APC750 &Cluster of Differentiation 16 - Antigen \\
        10&HLA-PB &Human Leukocyte Antigen \\
        11&CD45-KO & Cluster of Differentiation 45 - Antigen\\
        \bottomrule
    \end{tabular}}
\end{table}

\paragraph{Data and labels:}
Flow cytometry data is primarily tabular data.
Each sample is represented by a $N\times D$ matrix of values. It is therefore equivalent to a $D$-dimensional point cloud that cytometrists visualize via 2D projections. The result of such operations is the definition and quantification of different cell populations according to their phenotype. 

The dataset we propose contains samples from 30 patients;  each sample with an average of $\overline{N}=500'000$ cells (see supplementary material for the detailed count) and $D=12$ dimensions (markers, according to Table~\ref{tab:1}). At that stage, data has not undergone any other normalization than the one provided by the flow cytometry apparatus. 

While we acknowledge that our study utilizes a dataset with 12 markers, this number is well within the range of typical flow cytometry panels used in clinical practice. Many previous studies have successfully demonstrated the utility of machine learning techniques on similar or even lower dimensional flow data. For instance, \cite{Arvaniti2017} applied representation learning on a 10-marker dataset to detect rare disease populations. \cite{Bruggner2014}identified stratifying signatures across 13 surface markers, while \cite{Finak2009} proposed a mixture model-based approach for cell population identification on a datasets with 8-12 markers, showcasing the effectiveness of their method in accurately identifying cell populations across diverse marker panels. Moreover, the key challenge in FC analysis does not solely stem from high dimensionality but rather from the inherent heterogeneity, complexity, and rarity of cell populations of interest amidst a majority of irrelevant cells. Our dataset, comprising over 1 million cells per patient, with 5 and 6 distinct and unbalanced cell types, accurately reflects these real-world challenges faced by cytometrists.

\paragraph{Files:}
Files   have   been   	
anonymized and 
 stored in ``Flow Cytometry Standard'' (FCS)\footnote{\href{https://isac-net.org}{International Society for Advancement of Cytometry}} format that is generically produced by the flow cytometer.
Readers for such format include the {\tt FlowCal} Python library\footnote{\href{https://taborlab.github.io/FlowCal library}{Python {\tt FlowCal} library}} or the {\tt flowCore} package in the larger {\tt bioconductor} R library\footnote{\href{https://bioconductor.org/packages/release/bioc/html/flowCore.html}{R {\tt FlowCore} package}}.

%Files have been carefully investigated for anonymization in order to prevent medical information leakage.
Subsequently data files were cleaned manually with the help of the KALUZA software so that cell debris and dead cells were eliminated. The  populations of interest were then defined manually and stored as separate files for each patient. For each patient, we obtained therefore 
 seven data files: 
 \begin{itemize}
     \item 
\texttt{Case\_A.fcs} which contained the total cell population, 
\item \texttt{Case\_O.fcs} (first class) contained only T lymphocytes, \item \texttt{Case\_N.fcs} (second class) contained only B lymphocytes, \item \texttt{Case\_G.fcs} (third class) contained only monocytes, \item \texttt{Case\_P.fcs} (fourth class) contained only mast cells, \item \texttt{Case\_K.fcs} (fifth class) contained only hematopoietic (HSPCs) cells, 
\item \texttt{Case\_B.fcs} (sixth class) contained all the remaining cells denoted as $\left\{\text{B}\right\} = \left\{\text{A}\right\} \setminus \left\{\text{O},\text{N},\text{G},\text{P},\text{K}\right\}$.
 \end{itemize} Labels were kept aside and stored in basic ``Comma Separated Value'' (CSV) format that is easy to parse in any language.
This offers the possibility to add new labels to the same data.
The correspondence between cells (lines of FCS files) and their labels (lines in CSV files) is simply made by line numbering.

Overall, the data repository includes files for a total volume of around 1.3 GB.
The data is distributed in our convenient group repository\footnote{\href{https://viper.unige.ch/flowcyt}{FlowCyt Webpage}}.
%https://anonymous.4open.science/r/FlowCyt-Classification-Benchmark-4D53/README.md
\section{Structure of the Benchmark}
\label{sec:arch_benchmark}
FlowCyt aims to facilitate the flow cytometry process undertaken by hematologists by promoting reproducibility and extensibility in single-cell classification research. The benchmarking framework and dataset are openly accessible, allowing researchers to explore novel methodologies and algorithms while reproducing existing results. By providing a standardized evaluation framework and a richly annotated dataset, FlowCyt empowers researchers to push the boundaries of single-cell analysis in hematological/immunological flow cytometry data. 
%In collaboration with hematologists performing manual flow cytometry data analysis, we have identified the relevant tasks for automation.

\subsection{Classification Task:} \label{subsec:Classification_general} The main focus of our benchmark is to classify different cell types present in the heterogeneous samples. To make a fair comparison and evaluation of the performance of different baseline methods, we have provided various metrics, such as accuracy, precision, recall, and F1 score. Since it is a multi-class problem, these metrics are essential for ensuring robustness and comparability of the results. Moreover, being the dataset strongly imbalanced, we have computed the ratio of the correctly predicted labels for each class in each patient. These measures will assist researchers in precisely evaluating and comparing the classification models.

\subsubsection{Sub-Population Classification:} \label{subsec:sub_classification} We offer a specialized evaluation that focuses only on the classification of cells that belong to the $\left\{\text{O},\text{N},\text{G},\text{P},\text{K}\right\}$ population. This analysis allows researchers to examine the effectiveness of classification techniques in differentiating the variations within this specific subset of hematological cells. By isolating this distinct cell population, researchers can better understand the capacity of classification models to accurately identify and classify cells that are clinically relevant.

\subsubsection{Total Population Classification:} \label{subsec:total_classification} To evaluate the learning robustness and generalization capabilities of the classification models, FlowCyt includes a classification task that involves the entire population $A$ for each patient. By evaluating the models with the inherent complexity and diversity present within population $A$, researchers can determine how well the classification methodologies can work in real-world clinical scenarios.

\subsection{Other Tasks:} The benchmark we've created for flow cytometry data analysis goes beyond classification and can be used for various analytical tasks, such as clustering, dimensionality reduction, trajectory inference, and anomaly detection. The dataset can be explored by researchers to gain insights into cellular types, making it a versatile resource for researchers working on FCS data. FlowCyt is therefore a resource that we aim to enrich with new data, labels, and task definitions (see Section \ref{sec:FutureResearch}).\\
It will also de facto get enriched by the community via its usage.
 
\section{Experiments}
\label{sec:experiments}
We now discuss the details of our experimental design, which includes both inductive (IL) and transductive learning (TL) tasks. By combining these two learning methodologies, we aim to utilize their strengths to conduct a comprehensive analysis of our data. This approach also ensures that our results are more reliable and robust.

\subsection{Techniques}
To solve our tasks, which are mainly related to multi-class classification, we utilized well-known classification techniques as our baseline. All of these baselines were tested in the inductive learning framework, which was the basis for our medical evaluation, as depicted in Figure \ref{fig:test-k-fold-splitting} and fully described in Section \ref{subsect:IL}. Additionally, we tested the best-performing models in a transductive learning scenario to further evaluate the robustness of the learning process. This last scenario should be considered as an extension of the classification (inductive) task, where the reliability of the models was proven under semi-supervised learning, rather than full-supervised one.

\subsubsection{Baselines}
FlowCyt offers a variety of methods for single-cell classification, ranging from traditional techniques like Gaussian Mixture Models (GMM), XGBoost (XGB), and Random Forest (RF) to more advanced methods like Deep Neural Networks (DNNs) and Graph Neural Networks (GNNs). GNNs stand out in terms of performance, demonstrating the ability of graph learning techniques to identify patterns in a cell's neighborhood and their interactions. This enhances our understanding of hematological situations by utilizing spatial information and network structures. 

To use these graph learning methods, FlowCyt provides the option to convert tabular data to graph-structured data that can be fed into the various proposed GNNs. We achieved this by treating each cell as a node in the graph space, with $\mathcal{N}_i$ representing the set of $k$-nearest neighbors for node $i$ (where $k=7$) using the $\text{L}_2$ metric in the feature space. For each node, we built all possible pairs of its neighbors and added them to the edge set of the graph, resulting in a fully connected graph for each patient-graph instance. After experimenting with different values, we decided to use $k=7$ to strike a balance between making the graph too sparse (which could miss important connections) and too dense (which could include irrelevant connections). This approach enabled us to capture both the local structure (individual neighbors) and higher-order relationships (combinations of neighbors) in the data, helping us identify the most relevant relationships without adding noise. Moreover, constructing the graph via the $k$-nearest neighbor is intrinsically local and does not force us to make any assumptions about the distribution of the data, making it a good choice for datasets like ours where the full distribution of the data could be unknown.

\subsubsection{Inductive Learning}
\label{subsect:IL}
To accomplish this specific task, we used the dataset from our cohort of 30 patients. To ensure a reliable evaluation methodology, we employed a randomized approach. We carried out a 7-fold test procedure and we selected 10\% of each
training fold uniformly at random as a validation set to evaluate the predicted labels $\hat{y}_i$ by our model. This approach means that out of 30 patients, 4 or 5 are for testing, 2 or 3 for validation, and the rest for training. It's important to note that the testing graphs are completely unseen during training.

\begin{figure}[htbp]
\floatconts
  {fig:test-k-fold-splitting}
  {\caption{\footnotesize The dataset was split into seven segments for analysis. Six of these segments were used as training data, while one was reserved as test data. Additionally, 10\% of each training segment was randomly selected to create a validation set.}}
  {%
    \subfigure[\footnotesize 7-fold test procedure diagram]{\label{fig:circle}%
      \includegraphics[width=0.85\linewidth]{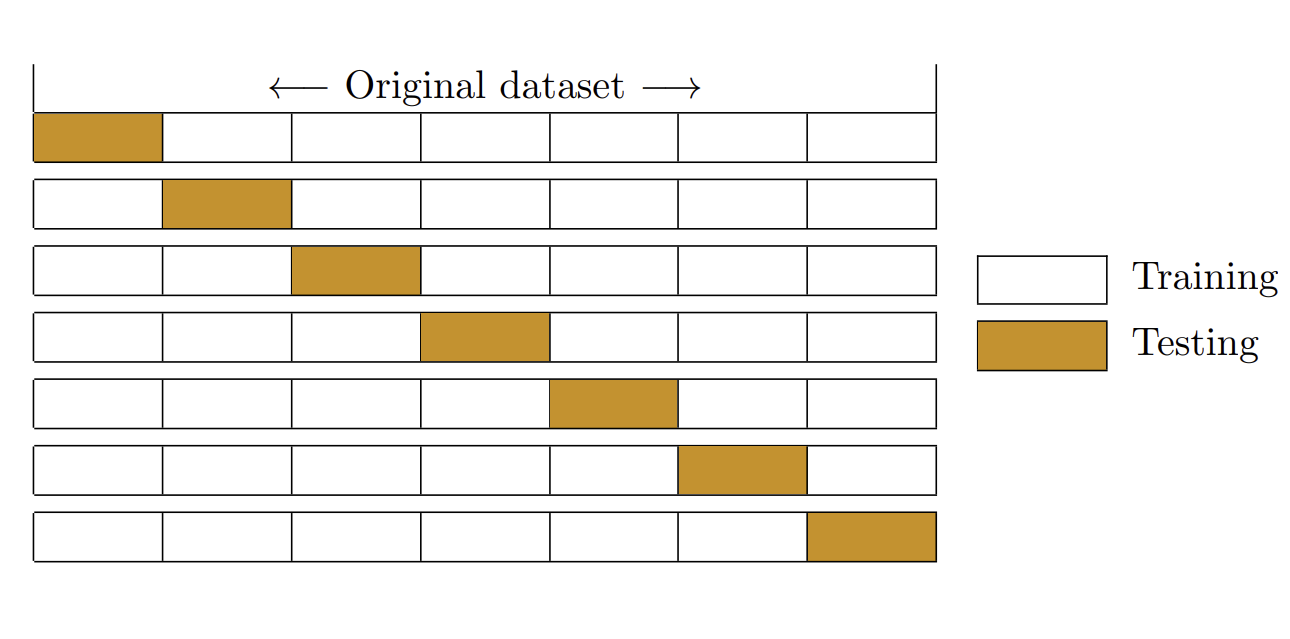}}%
    \qquad
    \subfigure[\footnotesize Train/Val splitting]{\label{fig:square}%
      \includegraphics[width=0.5\linewidth]{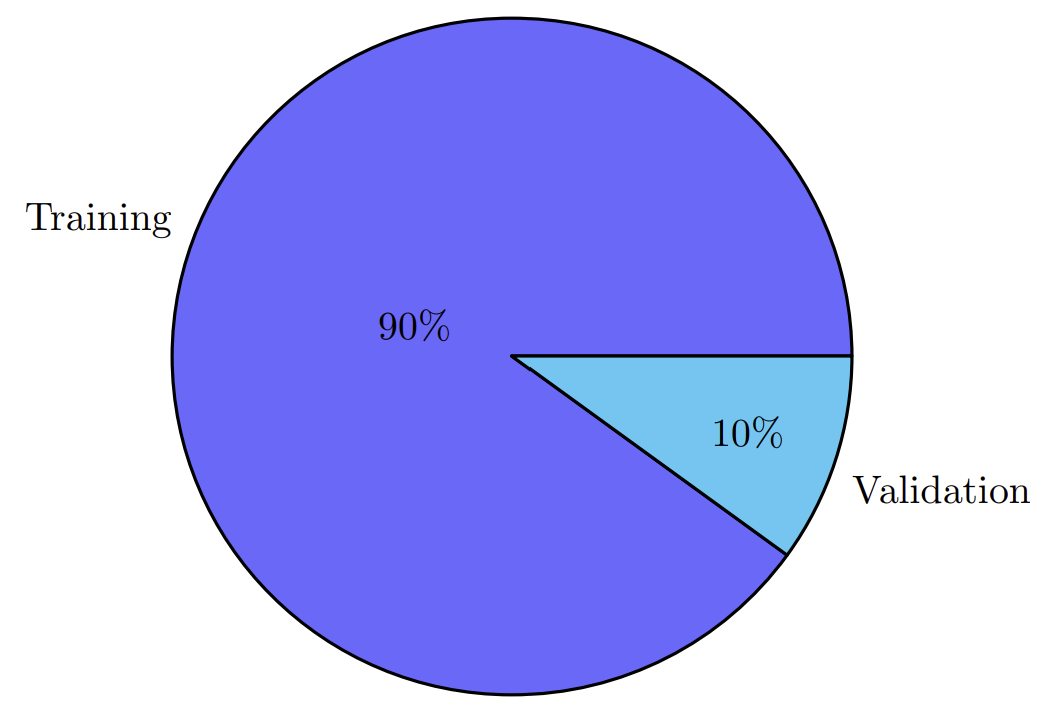}}
  }
\end{figure}

We trained our model and adjusted the hyperparameters on the validation set to assess performance on the test set. We ran our models five times with five different seed initialization to ensure the robustness of our results.
\paragraph{Transductive Learning}
\label{subsect:TL}
Motivated by the best results produced in Section \ref{subsect:IL}, we further proved the effectiveness of the models by using the transductive learning methodology outlined in \cite{yang2016}, where during the training phase the models were given access to the feature vectors of all nodes. While the concept of transductive learning has been explored in various domains, its application to flow cytometry data is relatively novel, and there is little literature explicitly discussing its implications in this field. \cite{Hu2020} and \cite{zhang2019scina} proposed a robust and interpretable end-to-end deep learning model for cytometry data and scRNA-Seq analysis, incorporating elements of self-supervised and semi-supervised learning. They discuss the advantages of leveraging unlabeled data to improve model generalization and robustness, which is crucial in the context of heterogeneous and high-dimensional cytometry data to reduce the need for extensive manual gating and annotation. 

In FlowCyt, our primary goal was to classify the status of cells, distinguishing between the five different classes, while also evaluating the reliability of the classification performances in a semi-supervised setup utilizing up to one million cells per patient. To achieve this, we conducted five different seed iterations using our models. We randomly assigned 10\% of the nodes for validation and 10\% for testing while keeping the classes' ratio balanced, the same as in the full dataset. Since our model is strongly imbalanced, we used a weighted negative log-likelihood loss function. Then we randomly masked the labels of 50\% of the training, validation, and test set. We can therefore think of our problem as having a large graph where half of the graph is unlabeled, and our goal is to predict the labels based on the ones that we have.

\subsection{Results}

In the following sections, we provide a detailed account of our experimental setup. We present the classification results for both sub and total cell populations (see Section \ref{subsec:Classification_general}) in the inductive scenario. Additionally, we present the TL results for the previous best models.

\subsubsection{Sub-Population IL}
We present here the results of the classification task as discussed in Section \ref{subsec:sub_classification}. We measure the performance in terms of accuracy, precision, recall, and F1 score as reported in Table \ref{tab:2}.
\begin{table}[hbtp]
    \footnotesize
    \centering
    \floatconts
    {tab:2}
    {\caption{\footnotesize Average metrics across all 30 patients on sub-population. All the results are averaged with $\pm 0.01$, and the names are shortened.}}
    {\begin{tabular}{ccccc}
    \toprule
        Model & Accuracy & Precision & Recall & F1-Score \\
        \midrule
        \textbf{GAT} & 0.97 & \textbf{0.99} & \textbf{0.98} & \textbf{0.98} \\
        SAGE & \textbf{0.98} & 0.97 & 0.98 & 0.98 \\
        GCN & 0.96 & 0.96 & 0.95 & 0.95 \\
        DNN & 0.86 & 0.88 & 0.84 & 0.80 \\
        RF & 0.94 & 0.91 & 0.83 & 0.88 \\
        XGB & 0.97 & 0.96 & 0.98 & 0.97 \\
        GMM & 0.47 & 0.77 & 0.36 & 0.50 \\
        \bottomrule
    \end{tabular}}
\end{table} We compare state-of-the-art methods for tabular data classification, such as DNNs, XGB, RF, and GMM, with popular GNNs like GraphSAGE (SAGE) \citep{SAGE}, Graph Convolutional Networks (GCNs) \citep{GCN}, and Graph Attention Networks (GATs) \citep{GAT}. 

We calculate these metrics for each class $i$, using the following formula (where TP stands for True Positive, FN for False Negative, and FP for False Positive):

\begin{equation}
  \begin{split}
    \text{Precision}_i &= \frac{\text{TP}_i}{\text{TP}_i + \text{FP}_i}, \\
    \text{Recall}_i &= \frac{\text{TP}_i}{\text{TP}_i + \text{FN}_i}, \\
    \text{F1 Score}_i &= 2 \times \frac{\text{Precision}_i \times \text{Recall}_i}{\text{Precision}_i + \text{Recall}_i}.
  \end{split}
\end{equation} We have similar equations for $i = 1, 2, 3, 4, 5$. Encouraged by the predictive ability of some of the models, we take a deeper look, and in Table \ref{tab:3a} and Table \ref{tab:3b}  we show the (correct) predicted label across patients and cell types.
\begin{table}[hbtp]
    \footnotesize
    \centering
    \floatconts
    {tab:3a}
    {\caption{\footnotesize Comparison of the average correct ratios across all 30 patients on the sub-population. Results are averaged with $\pm 0.01$.}}
    {\begin{tabular}{cccccc}
    \toprule
        Type & \textbf{GAT} & DNN & RF & XGB & GMM\\
        \midrule
        T Cells & \textbf{99.18} & 96.71 & 93.25 & 97.02 & 63.71 \\
        B Cells  & \textbf{90.53} & 73.62 & 90.60 & 82.63 & 51.00 \\
        Monocytes  & \textbf{97.89} & 90.28 & 90.38 & 93.01 & 25.06\\
        Mast  & \textbf{98.28} & 88.78 & 1.74 & 49.28 & 30.00 \\
        HSPC & \textbf{92.36} &  83.68 & 84.07 & 88.76 & 30.76\\
        \bottomrule
    \end{tabular}}
\end{table}

\begin{table}[hbtp]
    \footnotesize
    \centering
    \floatconts
    {tab:3b}
    {\caption{\footnotesize Cont'd of Table \ref{tab:3a} on the sup-population classification task.}}
    {\begin{tabular}{cccccc}
    \toprule
        Type & \textbf{GAT} & SAGE & GCN \\
        \midrule
        T Cells  & 99.10 & 98.18 & \textbf{99.12}\\
        B Cells  & \textbf{90.53} & 86.86 & 84.22\\
        Monocytes  & \textbf{97.19} & 93.71 & 96.55\\
        Mast Cells  & 98.28 & \textbf{99.50} & 97.24\\
        HSPC  & 92.36 & \textbf{93.25} & 92.14\\
        \bottomrule
    \end{tabular}}
\end{table}

\subsubsection{A-Population IL}
We summarise in Table \ref{tab:IL/A-pop} the results on the total cell population (Section \ref{subsec:total_classification}) for each patient, and we show in Table \ref{tab:hyperIL} of Appendix \ref{apd:first} the hyperparameters used throughout all the experiments. It is important to note that some models were not able to scale up on larger populations of up to one million nodes while maintaining a good ratio of correctly predicted labels. This indicates that only graph models, which are based on graph-encoded data, can capture the cell relations that are crucial for the classification task. While we only mentioned here GAT as a graph model, please refer to the code repository for all the results obtained using various GNNs related to Table \ref{tab:IL/A-pop}. Moreover, moved by the extremely accurate results provided by GAT, we also interpret the choices made by the model, analyzing which markers (features) were considered more important than others, as shown in Figure \ref{feature_importance} where the top 10 are shown.
\begin{figure}[htbp]
\footnotesize
\centering
\floatconts
  {feature_importance}
  {\caption{\footnotesize Feature importance as highlighted by the model. For the sake of simplicity, we remind the reader to match the labels with the corresponding Table \ref{tab:1} for future explanations.}}
  {\includegraphics[width=1.0\linewidth]{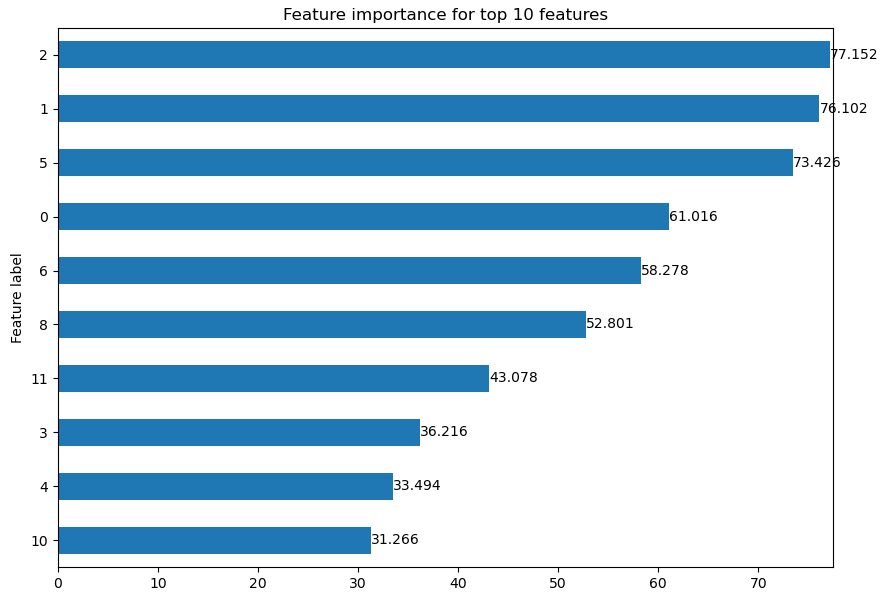}}
\end{figure} As we can see, the features that are given the most importance (at least above 55\%) are 2,1,5,0 and 6, therefore we break down the choice of the model for each one. In order of importance, we have:
\begin{itemize}
    \item CD14-FITC. The first important feature demonstrates  the expression level of CD14. CD14 is a cell surface receptor that binds to lipopolysaccharide (LPS), which is a component of the bacterial cell wall. CD14 is mainly expressed by monocytes, macrophages, and activated granulocytes. 
    It plays a role in the immune response against bacterial infections.
     A high importance of this feature may imply that our model can distinguish between cells that are involved in innate immunity and those that are not, detecting the presence of bacterial infection in the sample.
    \item Side Scatter (SSC)-Cell’s granularity. This feature denotes the level of light scatter at a 90-degree angle in relation to the laser beam. SSC reflects the internal complexity or granularity of the cell,  encompassing features like the presence of granules, nuclei, or other organelles. A high importance of this feature may suggest that either our model can differentiate between cells that have different degrees of complexity, such as lymphocytes, monocytes, and granulocytes, or  our model can identify cells that have abnormal granularity, such as blast cells or malignant cells.
    \item CD33-PC5.5. This feature indicates the expression level of CD33, a cell surface receptor that belongs to the sialic acid-binding immunoglobulin-like lectin (Siglec) family. CD33 is expressed by myeloid cells, such as monocytes, macrophages, granulocytes, and mast cells, and it modulates the immune response by inhibiting the activation of these cells. A high importance of this feature may suggest that our model can distinguish between myeloid and non-myeloid cells, or that our model can detect the expression of CD33 as a marker for certain types of leukemia, such as acute myeloid leukemia (AML) or chronic myelomonocytic leukemia (CMML).
    \item Forward scatter (FSC)-Cell’s size. This feature indicates the level of light scatter along the path of the laser beam. FSC is proportional to the diameter or surface area of the cell, and it can be used to discriminate cells by size. Placing emphasis on this feature may suggest that our model can differentiate between cells that have different sizes, such as small lymphocytes and large monocytes.
    \item CD34-PC7. This feature indicates the expression level of CD34, which is a cell surface glycoprotein that belongs to the sialomucin family. CD34 is expressed by HSPCs, and functions as a cell-cell adhesion molecule. The importance on this feature suggests that the model can identify the presence of HSPCs in the sample.
\end{itemize}

\begin{table}[hbtp]
    \footnotesize
    \centering
    \floatconts
    {tab:IL/A-pop}
    {\caption{\footnotesize Comparison of the average correct ratios across all 30 patients on the A-population. Results are averaged and rounded with $\pm 0.01$.}}
    {\begin{tabular}{cccccc}
    \toprule
        Type & \textbf{GAT} & DNN & RF & XGB & GMM\\
        \midrule
        T Cells & \textbf{97.91} & 91.86 & 62.67 & 81.45 & 72.04 \\
        B Cells  & \textbf{95.17} & 56.60 & - & 6.33 & - \\
        Monocytes  & \textbf{98.03} & 89.31 & - & 72.63 & 68.74\\
        Mast  & \textbf{97.40} & 47.09 & - &  15.54 & - \\
        HSPC & \textbf{90.79} &  32.79 & - & 24.89 & -\\
        Others & 87.77 &  \textbf{94.82} & 91.07 & 80.03 & -\\
        \bottomrule
    \end{tabular}}
\end{table}

\subsubsection{Populations on TL}
In this section, we discuss the results obtained for the sub-population on TL set-up (see Section \ref{subsec:sub_classification}). We have evaluated our metrics and found that the F1 score is around $\left\{0.94 \pm 0.01\right\}$ for GAT, GCN, and SAGE, which confirms the robustness of graph models and graph encoded data even for semi-supervised approach with large graphs. The results for the employed models have been obtained using those hyperparameters to fit the constraints of 1 GPU with 11GB memory (RTX 2080 Ti), as shown in Table \ref{tab:hyper_trans} of Appendix \ref{apd:first}. To highlight the strength of graph models, Figure \ref{fig:tSNE23} shows the t-SNE \citep{tSNE} embeddings visualization of GAT for one random patient (patient 23 has been randomly picked). Despite the difficulties due to the mask setup that is typical of transductive learning, we can see how the graph model has clustered the five classes into its final embedding layer. To view the results for the sub-population under the TL framework, please refer to our code repository.
\begin{figure}[htbp]
\centering
\floatconts
  {fig:tSNE23}
  {\caption{\footnotesize t-SNE projection for one random patient, for the transductive learning task.}}
  {\includegraphics[width=1.0\linewidth]{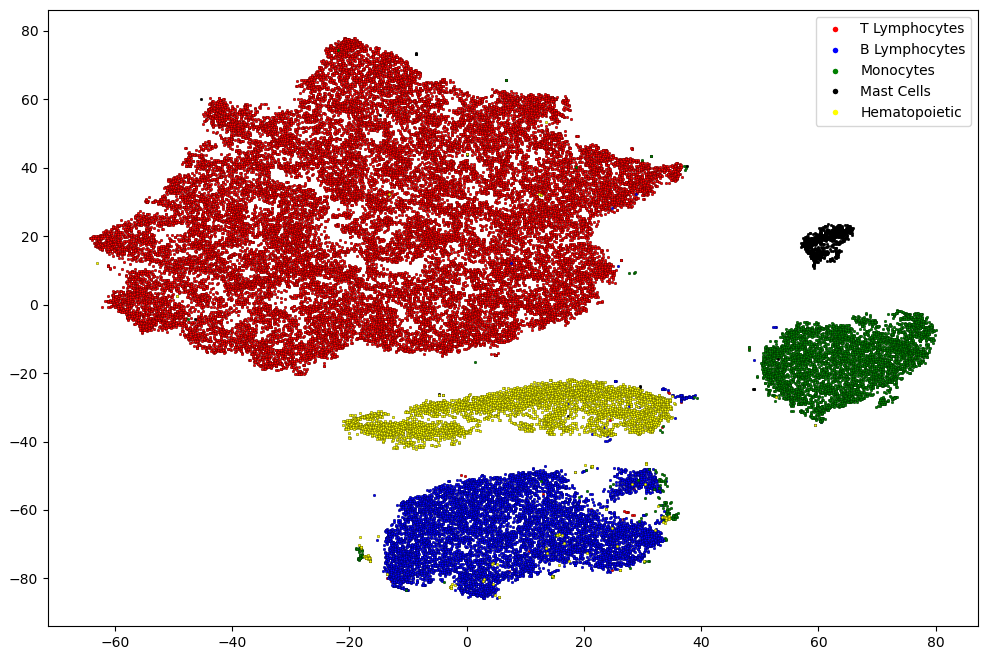}}
\end{figure}

\section{Discussion of FlowCyt}
\label{sec:discussion}
FlowCyt provides a valuable resource for researchers in the field of hematological cell populations. The dataset is extensively annotated and captures the complexity and heterogeneity of these cell populations. The high-dimensional tabular data allows for various analytical tasks, such as dimensionality reduction, clustering, and trajectory analysis, to be performed beyond classification. By utilizing their domain knowledge, researchers can explore the biological relationships and hierarchical structures among the cell types based on the provided labels. 

The feasibility of developing automated solutions to assist cytometrists has been demonstrated through multi-class classification experiments throughout Section \ref{sec:experiments}. The effectiveness of graph neural networks in modeling spatial patterns and inter-cell dependencies via graph encoding has been highlighted by the results obtained. This benchmark represents the first public flow cytometry benchmark that empowers rigorous assessment of novel methodologies. It will facilitate research that pushes the boundaries and clinical translation of automated single-cell analysis to supplement traditional manual examination. Through collaborative efforts, this versatile dataset may uncover additional tasks and biological insights, thus contributing to advancing the field of hematological cell populations \citep{hemagraph}.

\subsection{Future Research}
\label{sec:FutureResearch}
This benchmark is a solid foundation for expanding research in flow cytometry through diverse data and advanced analytical tasks. To better represent real clinical scenarios and improve model robustness, it is important to increase our patient cohort and include patients 
with hematologic/immunologic diseases, such as leukemias (e.g. Acute Myeloid Leukemia-AML).
%collect additional samples from healthy individuals and patients with various conditions, particularly malignant hematological disorders such as Acute Myeloid Leukemia (AML). 
Obtaining longitudinal samples that track patients over treatment courses could also reveal informative dynamics and could help to understand how cell populations evolve in response to therapies.

Advanced tasks should be defined in collaboration with hematologists/immunologists to ensure clinical relevance. There are opportunities to develop methodologies that support rather than replace manual analysis, providing a supplemental perspective. For example, automated strategies for quality control, across-patient normalization, batch effect correction, and data visualization could aid cytometrists. Beyond improving existing practices, innovative new paradigms could extract novel biological insights. 

Moreover, there are numerous learning paradigms for flow cytometry: hierarchical relationships between hematopoietic cells could be modeled \citep{leukograph}, 
aligning with developmental pathways to identify pathological deviations from healthy cell distribution patterns. Hybrid approaches could combine unsupervised representation learning with supervised diagnosis. 

Another promising direction is single-cell trajectory inference (TI), where apart from providing insights into phenotypes of different blood cell types, FlowCyt's dataset can also be used for better understanding the developmental trajectories of hematopoietic precursors. The use of our dataset is not limited to supervised classification problems and can be extended to benchmarking different trajectory inference methods, which so far have been reviewed for transcriptomics \citep{TrajInference} and not flow cytometry data. In such a case, access to cell type annotations will be valuable to evaluate the quality of the reconstructed trajectories and compare the performance of different TI methods suitable for non-time resolved data. Moreover, our cohort data will allow us to establish a healthy trajectory model that will serve in the future as a reference to comprehend in more detail the differences between healthy and malignant hematopoiesis.

\section{Conclusion}
Overall, this benchmark enables diverse analytical approaches from conventional supervised learning to emerging paradigms in flow cytometry research. Collaborative efforts to generate new labeled datasets, formulate meaningful clinical questions, and develop inventive computational methodologies will catalyze impactful discoveries in hematology and immunology. By fostering rigorous yet creative investigations leveraging its versatility, this benchmark aims to accelerate the advancement of automated single-cell analysis.

\section*{Acknowledgments}
The Swiss National Science Foundation partially funds this work under grant number 207509 "Structural Intrinsic Dimensionality".

%\newpage
\bibliography{flowcyt}

\appendix

\section{Hyperparameters}
\label{apd:first}
Tables \ref{tab:hyper_trans} and Table \ref{tab:hyperIL} below show the default hyperparameters used in all the experiments:

\begin{table}[hbtp]
    \centering
    \tiny
    \floatconts
    {tab:hyperIL}
    {\caption{\footnotesize Default hyperparameters used in all inductive experiments. For XGBoost, Random Forest, and Gaussian Mixture Model we used the sklearn package as shown in our code repository.}}
    {\begin{tabular}{ccccc}
    \toprule
        Params & GAT & SAGE & GCN & DNN\\
        \midrule
        Layers  &3 & 3 & 3 &3 \\
        Hidden-ch  &64 &64 & 64 &256\\
        Att-heads &8 &- & -  &-\\
        Optimizer &Adam &Adam & Adam  &Adam\\
        Lr-sched  & [0.01,1e-7] &[0.01,1e-7] &[0.01,1e-7] &[0.01,1e-7]\\
        Weigh-decay &0.0005 &0.005 & 0.005 &0.005\\
        Dropout  &0.3 &0.4 & 0.4  &0.5\\
        Epochs  &1000 &1000 & 1000  &1000\\
        Early-stop  &40 &50 & 50  &40\\
        \bottomrule
    \end{tabular}}
\end{table}
\begin{table}[hbtp]
    \footnotesize
    \centering
    \floatconts
    {tab:hyper_trans}
    {\caption{\footnotesize Default hyperparameters used in all transductive experiments to fit our memory GPU constraints.}}
    {\begin{tabular}{cccc}
    \toprule
        Parameter & GAT & SAGE & GCN\\
        \midrule
        Layers  & 3 & 3 & 3\\
        Hidden-ch  & 8 & 16 & 16\\
        Att-heads & 8 & - & -\\
        Optimizer & Adam & Adam & Adam\\
        Lr scheduler  & [0.01,1e-7] & [0.01,1e-7] & -\\
        Weight decay & 0.0005 & 0.005 & 0.005\\
        Dropout  & 0.3 & 0.5 & 0.5\\
        Train epochs  & 1000 & 1000 & 1000\\
        Early stopping  & 40 & 50 & 50\\
        \bottomrule
    \end{tabular}}
\end{table}

\end{document}